\documentclass[11pt]{article}

\usepackage[final]{acl}

\usepackage{times}
\usepackage{latexsym}

\usepackage[T1]{fontenc}

\usepackage[utf8]{inputenc}

\usepackage{microtype}

\usepackage{inconsolata}

\usepackage{graphicx}

\usepackage{algorithm}
\usepackage{fontawesome}
\usepackage{algorithmic}
\usepackage{times}
\usepackage{latexsym}
\usepackage{booktabs}
\usepackage{colortbl}
\usepackage{xcolor}
\usepackage{subcaption}

\definecolor{headercolor}{RGB}{230,230,250}
\definecolor{rowcolor1}{RGB}{245,245,250}
\definecolor{rowcolor2}{RGB}{255,255,255}
\definecolor{bestcolor}{RGB}{220,255,220}
\definecolor{headerblue}{RGB}{100, 100, 220}
\definecolor{lightblue}{RGB}{198, 219, 239}
\definecolor{altrow}{RGB}{240, 248, 255}
\usepackage[most]{tcolorbox}

\definecolor{oddoneout}{RGB}{255,182,193}
\definecolor{oddoneoutlight}{RGB}{255,182,193} 
\definecolor{manufacturing}{RGB}{30,144,255}    
\definecolor{manufacturinglight}{RGB}{173,216,230} 
\definecolor{material}{RGB}{34,139,34}         
\definecolor{materiallight}{RGB}{152,251,152}  
\definecolor{grouping}{RGB}{255,140,0}         
\definecolor{groupinglight}{RGB}{255,218,185}  
\definecolor{chronological}{RGB}{138,43,226}   
\definecolor{chronologicallight}{RGB}{221,160,221} 
\definecolor{attribution}{RGB}{184,134,11}     
\definecolor{attributionlight}{RGB}{255,255,224} 

\definecolor{example}{RGB}{25,25,112}          
\definecolor{examplelight}{RGB}{240,248,255}   
\definecolor{headercolor}{RGB}{230,230,250}
\definecolor{rowcolor1}{RGB}{245,245,250}
\definecolor{rowcolor2}{RGB}{255,255,255}
\definecolor{bestcolor}{RGB}{220,255,220}
\definecolor{headerblue}{RGB}{100,100,220}
\definecolor{lightblue}{RGB}{198,219,239}
\definecolor{altrow}{RGB}{240,248,255}
\definecolor{correct}{RGB}{0,128,0}            
\definecolor{correctlight}{RGB}{144,238,144}   
\definecolor{incorrect}{RGB}{220,20,60}    
\definecolor{incorrectlight}{RGB}{255,182,193} 
\definecolor{questionbg}{RGB}{240,248,255}
\definecolor{taskblue}{RGB}{25,25,112}         

\title{On the Cultural Anachronism and Temporal Reasoning \\ in Vision Language Models}

\author{
    {\bfseries Mukul Ranjan$^{1}$}\quad
    {\bfseries Prince Jha$^{1}$}\quad
    {\bfseries Khushboo Kumari$^{2}$}\quad
    {\bfseries Zhiqiang Shen$^{1}$} \\
    { $^{1}$MBZUAI, UAE}\quad
    { $^{2}$Inception, UAE}\quad
    \vspace{10pt}\\
    \faGlobe~\url{https://khushboo0012.github.io/tab-vlm-webpage/}\\
}

\begin{document}
\maketitle
\begin{abstract}
Vision-Language Models (VLMs) are increasingly applied to cultural heritage materials, from digital archives to educational platforms. This work identifies a fundamental issue in how these models interpret historical artifacts. We define this phenomenon as \textit{cultural anachronism}, the tendency to misinterpret historical objects using temporally inappropriate concepts, materials, or cultural frameworks. To quantify this phenomenon, we introduce the Temporal Anachronism Benchmark for Vision-Language Models (\textbf{TAB-VLM}), a dataset of 600 questions across six categories, designed to evaluate temporal reasoning on 1,600 Indian cultural artifacts spanning prehistoric to modern periods. Systematic evaluations of ten state-of-the-art models reveal significant deficiencies on our benchmark, and even the best model (GPT-5.2) achieves only 58.7\% overall accuracy. The performance gap persists across varying architectures and scales, suggesting that cultural anachronism represents a significant limitation in visual AI systems, regardless of model size. These findings highlight the disparity between current VLM capabilities and the requirements for accurately interpreting cultural heritage materials, particularly for non-Western visual cultures underrepresented in training data. Our benchmark provides a foundation for enhancing temporal cognition in multimodal AI systems that interact with historical artifacts. The dataset and code are available in our project page.
\end{abstract}

\section{Introduction}
\label{sec:intro}

Vision Language Models (VLMs) have demonstrated impressive capabilities in understanding visual content across diverse domains, from natural scene analysis to medical imaging \cite{radford2021learning, yu2022coca, liu2023llavavisual,liu2024improvedllava,bai2025qwen2.5vl,li2025benchmark,liu2025culturevlm}. Recently, their integration into cultural heritage applications has accelerated rapidly, with deployments spanning digital museum collections, educational platforms, and automated cataloging systems \cite{li2024culturellm, hwang2025cats,trichopoulos2023large,arnold2024explainable,ghaboura2025time}. However, as these models increasingly mediate between the public and cultural artifacts, we identify a critical gap: their tendency to interpret historical artifacts through inappropriate temporal lenses, a phenomenon we define as \textbf{\texttt{cultural anachronism}}.

Cultural anachronism in VLMs represents the systematic misattribution of concepts, techniques, materials, or interpretive frameworks to artifacts from time periods where they did not exist or were not culturally relevant. For example, when a VLM describes a sculpture from the 3rd century BCE using artistic vocabulary that emerged in 19th century European contexts, attributes manufacturing techniques to ancient pottery that were only developed millennia later, or interprets religious iconography through contemporary rather than period-appropriate symbolic frameworks \cite{liang2022holistic, bhatia2024datelogicqa,gallegos2024bias,ko2023large}. Such temporal-conceptual displacement poses a significant challenges for applications in museum digitization, educational technologies, and cultural heritage preservation, where accuracy in historical representation is paramount \cite{siliutina2024cultural,hwang2025cats,trichopoulos2023large,arnold2024explainable}. Unlike general object recognition errors that may cause minor inconveniences, anachronistic interpretations fundamentally misrepresent the cultural and historical significance of artifacts, potentially reinforcing colonial or contemporary biases in historical narratives and undermining the integrity of cultural heritage documentation \cite{birhane2021multimodal, thylstrup2022ethics}.

The challenge of temporal reasoning in VLMs extends beyond simple date recognition to encompass complex understanding of technological evolution, artistic development, and cultural context across historical periods. While existing benchmarks evaluate VLMs on contemporary visual understanding tasks \cite{li2025benchmark, chang2024survey}, none systematically assess their ability to maintain temporal coherence when interpreting historical materials. This gap is particularly substantial given the increasing reliance on AI systems for cultural heritage applications, where the stakes of misinterpretation extend far beyond technical performance metrics to encompass cultural representation, educational accuracy, and preservation of historical knowledge. The problem is compounded by the fact that VLMs are predominantly trained on contemporary visual data, creating inherent biases toward modern interpretive frameworks when encountering historical artifacts.

\begin{figure*}[!h]
    \centering
    \includegraphics[width=0.80\textwidth]{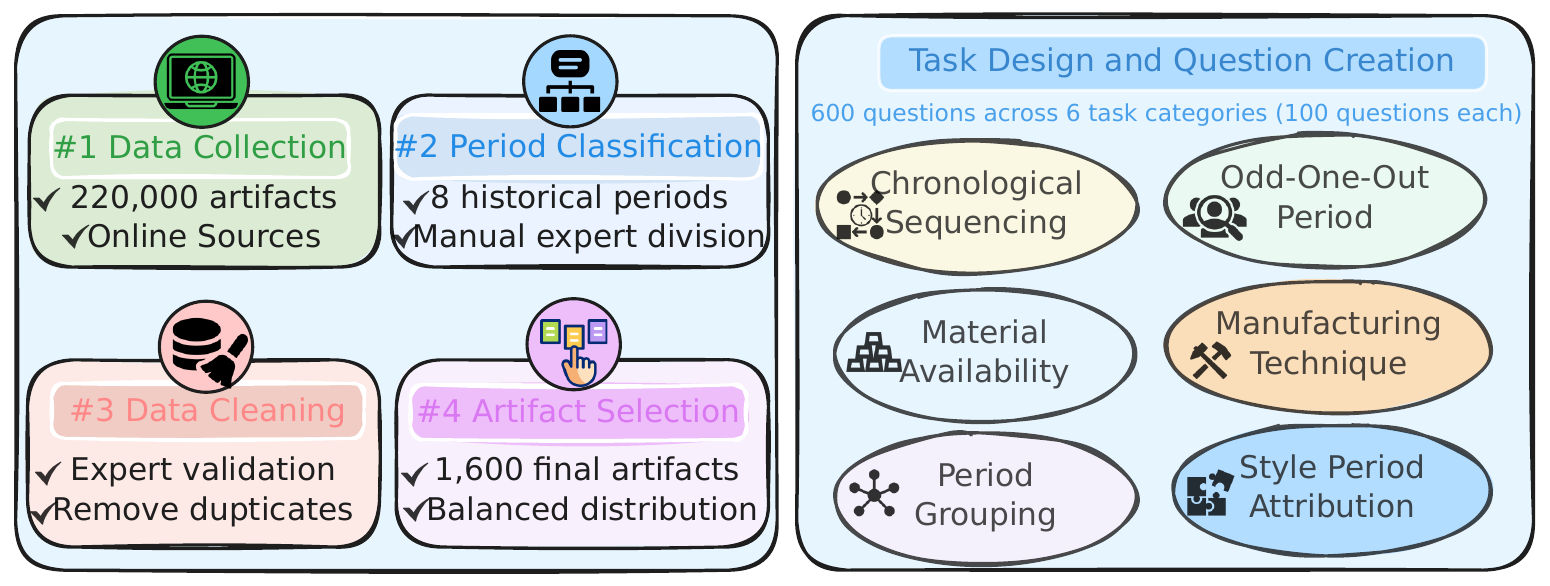}
    \caption{Data collection and processing pipeline showing systematic reduction from 220,000 initial artifacts to 1,600 curated items through expert validation and balanced selection.}
    \label{fig:data_pipeline}
\end{figure*}

To address this gap, we introduce the Temporal Anachronism Benchmark for Vision–Language Models (\textbf{TAB-VLM}), a comprehensive dataset comprising 600 carefully curated questions across six distinct evaluation categories, utilizing 1,600 artifacts spanning prehistoric to modern Indian history. Figure \ref{fig:data_pipeline} provides an overview of the artifact collection and curation pipeline used to construct TAB-VLM. Our benchmark specifically targets the temporal reasoning capabilities required for accurate historical artifact interpretation, including period attribution, chronological sequencing, anachronism detection, manufacturing technique identification, material availability assessment, and cultural context understanding. The selection of Indian cultural heritage as our focus domain provides several advantages: it encompasses a vast temporal range from prehistoric Indus Valley artifacts to contemporary works, represents diverse artistic traditions and technological developments, and offers rich documentation that enables precise temporal categorization while highlighting the global importance of non-Western cultural heritage in AI evaluation frameworks.

Our systematic evaluation approach addresses three fundamental research questions that illuminate the scope and nature of cultural anachronism in current VLMs:

\begin{enumerate}
    \item \textbf{RQ1:} To what extent do current state-of-the-art VLMs exhibit cultural anachronism when interpreting historical artifacts, and does this vary across models of different architectures and scales?
    
    \item \textbf{RQ2:} Which aspects of temporal reasoning (e.g., chronological ordering, manufacturing technique identification, material appropriateness) present the greatest challenges for VLMs, and what patterns emerge in their anachronistic interpretations?
    
    \item \textbf{RQ3:} How does performance on temporal reasoning tasks correlate with general visual understanding abilities, and what implications does this have for the development of VLMs that avoid cultural anachronism?
\end{enumerate}

We evaluate ten state-of-the-art models including both proprietary models (GPT-5.2~\citep{openai2025gpt5.2}, GPT-4o~\citep{hurst2024gpt4o}, GPT-4o-mini~\citep{hurst2024gpt4o}) and leading open-source alternatives (Qwen2-VL~\citep{wang2024qwen2vl}, Qwen2.5-VL~\citep{bai2025qwen2.5vl}, InternVL3~\citep{zhu2025internvl3} series) on our benchmark, and provide the first quantitative characterization of cultural anachronism prevalence and patterns in VLMs. Our findings reveal limitations in the historical reasoning capabilities of current models despite their demonstrated proficiency in general visual understanding tasks, with implications extending beyond technical performance to encompass broader questions of cultural competence, responsible AI deployment, and the preservation of accurate historical knowledge in digital contexts. This work establishes cultural anachronism as a notable evaluation dimension for multimodal AI systems and provides a foundation for developing more temporally-aware and culturally sensitive VLMs.

\vspace{-0.1in}
\section{Related Work}
\vspace{-0.1in}
Temporal reasoning remains a persistent challenge for AI systems, with benchmarks like TRAM~\citep{wang2023tram}, Test-of-Time~\citep{fatemi2024test}, TemporalVQA~\citep{imam2025can}, SpookyBench~\citep{upadhyay2025time} and TimeBench~\citep{chu2023timebench} revealing consistent deficiencies in time-based understanding despite advances in model architecture~\citep{jain2023language}. Recent approaches including self-critique methods \citep{su2024timo}, explainable frameworks \citep{yuan2024back}, video-based temporal reasoning~\citep{ko2023large, chen2024rextime}, and multimodal temporal-causal evaluation \citep{wang2025timecausality, padlewski2024vibeeval} show only modest improvements, with models particularly struggling to maintain period-appropriate perspectives \citep{underwood2025can}. Cultural understanding presents parallel challenges, with research revealing significant gaps in models' cultural knowledge \citep{li2024culturellm, rao2024normad, chiu2024culturalbench, liu2023multilingual, hwang2023aligning}, particularly for non-Western traditions and across temporal contexts. Multimodal evaluations \citep{nayak2024benchmarking, liu2025culturevlm, kannen2024beyond} demonstrate frequent cultural inaccuracies in VLMs, while frameworks for cross-cultural alignment \citep{kharchenko2024well, li2024culturepark, fung2024massively} remain limited in addressing temporal evolution of cultural expressions.
\vspace{-0.1in}

The intersection of temporal and cultural reasoning becomes important in cultural heritage applications, where AI systems must navigate both historical context and cultural representation. While classification systems for cultural heritage \citep{hwang2025cats} and heritage search interfaces \citep{trichopoulos2023large, arnold2024explainable} show promise, they struggle with temporal nuance and dynamic cultural evolution \citep{adilazuarda2024towards}. Recent benchmarks for historical artifacts \citep{ghaboura2025time} begin addressing these challenges, but they do not explicitly operationalize cultural–temporal reasoning as an evaluation target, nor do they expose anachronistic failure modes tied to historical artifacts. Temporal reasoning capabilities require deeper examination given the importance of accurate representation in digital cultural preservation \citep{siliutina2024cultural}.  These challenges connect to broader representational biases \citep{birhane2021multimodal, thylstrup2022ethics, gallegos2024bias} that particularly affect cultural representation, necessitating context-specific approaches \citep{sambasivan2021re} for artifacts potentially underrepresented in training data. Existing surveys~\citep{li2025benchmark, chang2024survey} document evaluation approaches for VLMs, but do not capture this dimension, motivating the need for targeted benchmarks such as TAB-VLM.

\begin{figure*}[t]
    \centering
    \includegraphics[width=0.85\textwidth]{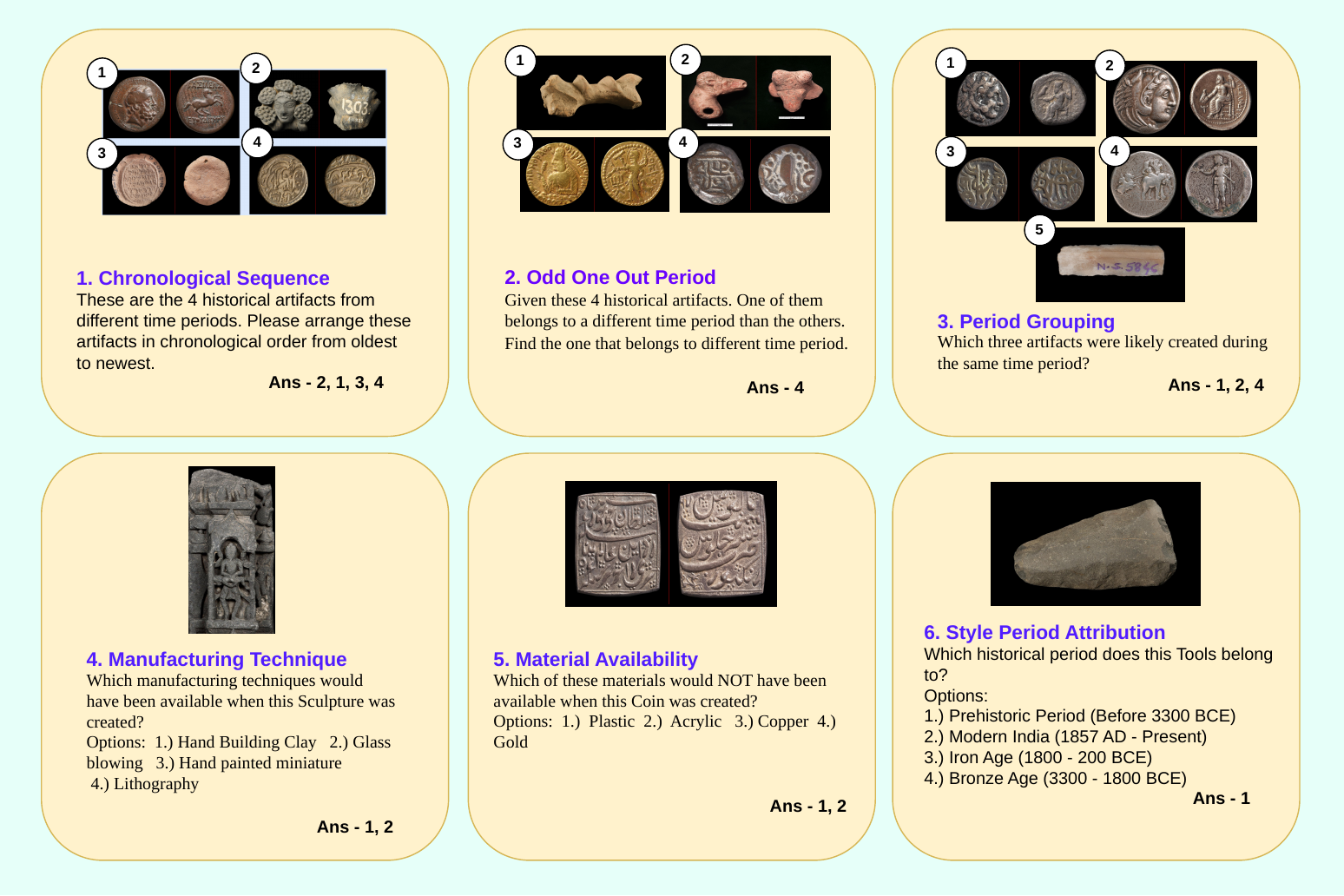}
    \caption{Examples of the six task types in our benchmark: chronological sequencing, odd-one-out period detection, material availability, manufacturing technique, period grouping, and style-period attribution.}
    \label{fig:task_examples}
\end{figure*}

\begin{figure}[htbp]
    \centering
    \includegraphics[width=0.85\linewidth]{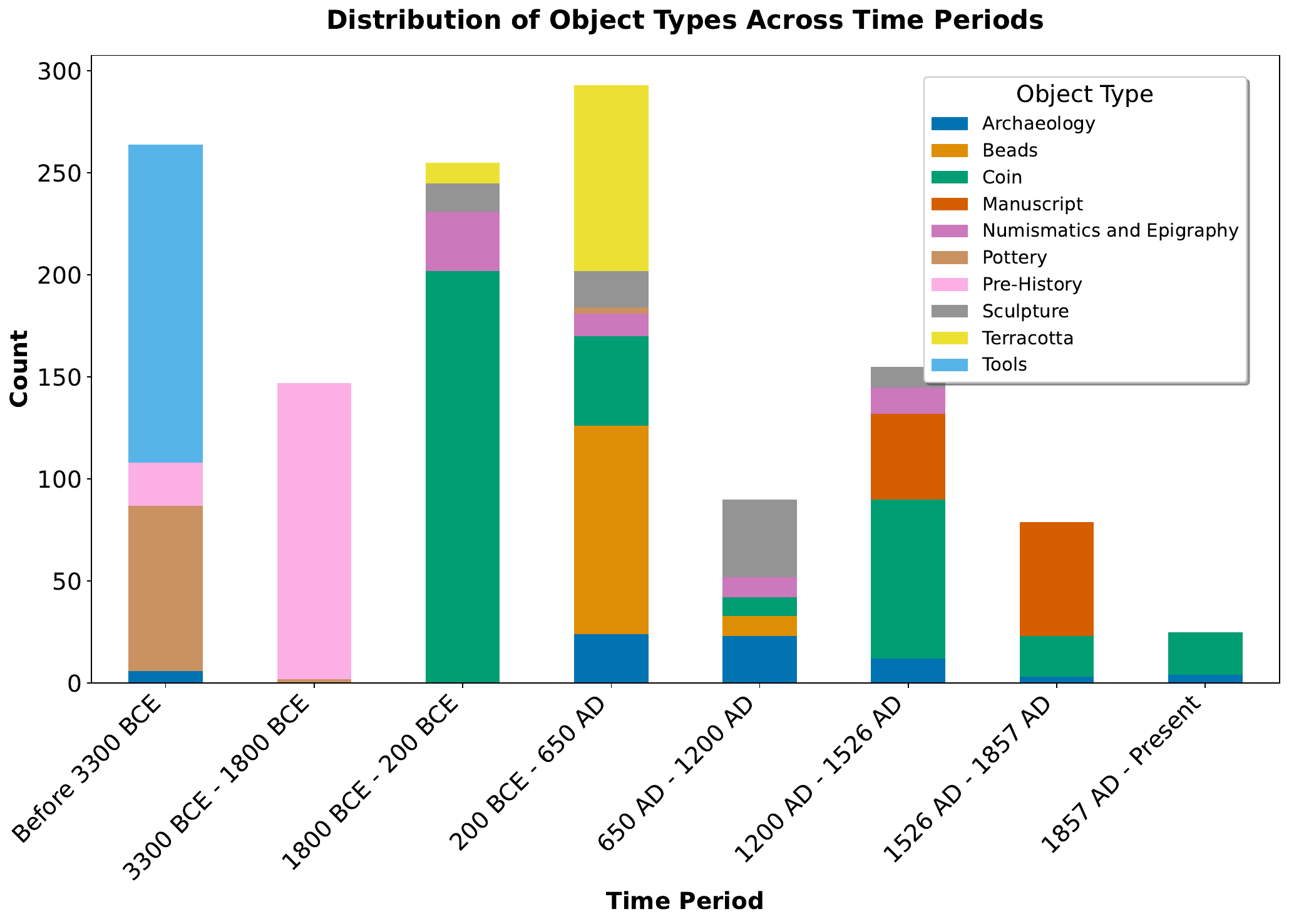}
    \caption{Distribution of object types across time periods indicating prevalence of terracotta artifacts in prehistoric eras and manuscripts during the Early Modern period.}

    \label{fig:object_types}
    \vspace{-0.2in}
\end{figure}


\section{TAB-VLM}

Our benchmark TAB-VLM focuses on Indian cultural heritage. We curate a diverse dataset of artifacts and construct temporal reasoning tasks that uncover anachronistic responses across six dimensions: (1) Chronological Sequence, (2) Period Intrusion Detection (Odd-One-Out Period), (3) Material Availability, (4) Manufacturing Technique, (5) Period-Based Grouping, and (6) Style Period Attribution. TAB-VLM offers a structured framework for assessing historical awareness in multimodal models.

\begin{table*}[t]
\centering
\resizebox{\textwidth}{!}{%
\begin{tabular}{lccccccc}
\toprule
\textbf{Model} & 
\textbf{Overall} & 
\textbf{Style-Period} & 
\textbf{Chronological} & 
\textbf{Manufacturing} & 
\textbf{Material} & 
\textbf{Period} & 
\textbf{Odd-One-Out} \\

& 
\textbf{Accuracy} & 
\textbf{Attribution} & 
\textbf{Sequence} & 
\textbf{Technique} & 
\textbf{Availability} & 
\textbf{Grouping} & 
\textbf{Period} \\
\midrule
\multicolumn{8}{l}{\textit{Proprietary Models}} \\
\midrule
GPT-5.2 & 
\textbf{58.7 $\pm$ 1.3} & 
\textbf{65.0 $\pm$ 4.1} & 
\textbf{37.2 $\pm$ 0.6} & 
\textbf{56.0 $\pm$ 1.2} & 
\textbf{92.1 $\pm$ 1.3} & 
\textbf{45.2 $\pm$ 1.3} & 
\textbf{57.1 $\pm$ 1.2} \\

GPT-5-mini & 
51.2 $\pm$ 0.4 & 
61.0 $\pm$ 0.3 & 
32.3 $\pm$ 2.1 & 
52.1 $\pm$ 1.1 & 
88.0 $\pm$ 1.2 & 
36.2 $\pm$ 0.8 & 
49.1 $\pm$ 1.0 \\

GPT-4o & 
50.4 $\pm$ 0.2 & 
60.3 $\pm$ 1.2 & 
30.7 $\pm$ 0.0 & 
49.8 $\pm$ 1.5 & 
85.3 $\pm$ 0.9 & 
32.3 $\pm$ 1.3 & 
45.3 $\pm$ 1.2 \\

GPT-4o-mini & 
47.2 $\pm$ 0.7 & 
45.0 $\pm$ 1.5 & 
34.0 $\pm$ 0.7 & 
39.0 $\pm$ 1.2 & 
89.0 $\pm$ 1.3 & 
28.0 $\pm$ 1.5 & 
48.0 $\pm$ 0.6 \\

\midrule
\multicolumn{8}{l}{\textit{Open-Source Models}} \\
\midrule
Qwen2-VL-7B & 
32.8 $\pm$ 1.4 & 
52.7 $\pm$ 0.9 & 
16.7 $\pm$ 0.2 & 
16.7 $\pm$ 1.3 & 
70.7 $\pm$ 0.6 & 
10.0 $\pm$ 1.2 & 
29.3 $\pm$ 1.2 \\

Qwen2.5-VL-7B & 
42.6 $\pm$ 1.1 & 
58.8 $\pm$ 0.4 & 
14.0 $\pm$ 0.3 & 
41.5 $\pm$ 0.9 & 
69.0 $\pm$ 0.4 & 
22.0 $\pm$ 0.6 & 
51.0 $\pm$ 0.8 \\

Qwen2.5-VL-3B & 
29.0 $\pm$ 2.1 & 
52.5 $\pm$ 1.1 & 
11.5 $\pm$ 1.3 & 
40.0 $\pm$ 1.7 & 
25.4 $\pm$ 0.6 & 
13.3 $\pm$ 1.8 & 
30.4 $\pm$ 1.1 \\

Qwen2-VL-2B & 
19.7 $\pm$ 0.5 & 
47.5 $\pm$ 0.5 & 
1.3 $\pm$ 1.8 & 
10.5 $\pm$ 0.7 & 
15.2 $\pm$ 0.2 & 
8.1 $\pm$ 1.3 & 
25.7 $\pm$ 0.4 \\

InternVL3-8B & 
36.2 $\pm$ 1.5 & 
53.0 $\pm$ 0.4 & 
9.0 $\pm$ 0.6 & 
41.0 $\pm$ 1.3 & 
59.0 $\pm$ 0.3 & 
16.0 $\pm$ 0.2 & 
39.0 $\pm$ 0.4 \\

InternVL3-2B & 
30.5 $\pm$ 0.1 & 
52.0 $\pm$ 0.5 & 
8.0 $\pm$ 0.5 & 
42.0 $\pm$ 0.6 & 
45.0 $\pm$ 1.3 & 
8.0 $\pm$ 0.1 & 
28.0 $\pm$ 1.2 \\

\midrule
\multicolumn{8}{l}{\textit{Baseline}} \\
\midrule
Random & 
12.8 & 
25.0 & 
4.2 & 
6.25 & 
6.25 & 
10.0 & 
25.0 \\

\bottomrule
\end{tabular}%
}
\caption{Performance comparison of VLMs on TAB-VLM benchmark with 100 questions per category. All values represent percentages (mean $\pm$ standard deviation across five runs). Best results in bold.}
\label{tab:vlm_comparison_updated}
\end{table*}

\subsection{Task Design and Evaluation Framework}

We construct 600 multiple-choice questions distributed equally across six different temporal reasoning task categories, with 100 questions per category, for the TAB-VLM benchmark. We construct each question using task-specific natural language prompt templates (provided in Appendix \ref{sec:prompt_template}), paired with one or more artifact images. We now describe each of the six task categories in detail.

\noindent\textbf{Chronological Sequencing.} The chronological sequencing task evaluates fundamental temporal ordering capabilities by requiring models to arrange artifacts from different periods in correct chronological sequence. Prompt template: "<image1>, <image2>, <image3>, <image4> These are the 4 historical artifacts from different time periods. Please arrange these artifacts in chronological order from oldest to newest." \textit{Task structure:} 4 artifacts with one correct ordering from 24 possible permutations.

\noindent\textbf{Odd-One-Out Period.} This task assesses a model's sensitivity to temporal anomalies by presenting a group of artifacts, one of which belongs to a different historical period than the others. The model must identify the artifact that does not fit, based on subtle visual cues that indicate differences in era or context. Prompt template: "<image1>, <image2>, <image3>, <image4>, Given these 4 historical artifacts one of them belongs to a different time period than the others. Find the one that belongs to different time period." \textit{Task structure:} 4 options with one correct answer.

\noindent\textbf{Material Availability.} This task evaluates a model's understanding of historical material by asking it to identify materials that would not have been available during the artifact's period. It tests knowledge of technological timelines to detect anachronistic material associations. Prompt template: "<image>. Which of the following materials would NOT have been available when this artifact was created?" \textit{Task structure:} 4 material options with 1-2 anachronistic materials (exact set match required).

\noindent\textbf{Manufacturing Technique.} This task tests a model's understanding of historical production methods by asking which manufacturing techniques would have been available during the time the artifact was created. It evaluates temporal knowledge of technological capabilities and craftsmanship. Prompt template: "<image>. Which manufacturing techniques would have been available when this artifact was created?" \textit{Task structure:} 4 technique options with 1-2 anachronistic techniques (exact set match required).

\noindent\textbf{Period Grouping.} This task examines pattern recognition across stylistic and material elements by requiring identification of contemporaneous artifacts, evaluating models' ability to recognize coherent aesthetic traditions and technological signatures characteristic of specific historical periods. Prompt template: "<image1>, <image2>, <image3>, <image4>, <image5> Which three artifacts were likely created during the same time period?" \textit{Task structure:} 5 artifacts where 3 belong to the same period (one correct combination from 10 possible).

\noindent\textbf{Style-Period Attribution.} This task provides assessment of temporal classification capabilities by requiring models to match artifacts with their correct historical periods from multiple options. It tests models' understanding of  visual characteristics, artistic conventions, and cultural expressions associated with specific temporal periods. Prompt template: "<image> Which historical period does this artifact belongs to?" \textit{Task structure:} 4 period options with one correct answer.

In all the prompt templates above, <image> denotes the visual input of a specific artifact. \textbf{Modality clarification:} Models receive only visual input (artifact images) during evaluation; no textual metadata about artifacts is provided. Figure~\ref{fig:task_examples} illustrates examples for each task, showing the image, question, and answer triplet. Random baseline performance for each task type is reported in Table~\ref{tab:vlm_comparison_updated}.

\subsection{Dataset Construction and Curation}

Our dataset was constructed through a rigorous four-stage pipeline, beginning with the collection of approximately 220,000 artifacts from online repositories~\cite{museumsofindia2025}. As shown in Table~\ref{tab:period_distribution} these artifacts were then classified into eight distinct historical periods, based on the taxonomy outlined in~\cite{mcleod2015history}: Prehistoric Period (before 3300 BCE), Bronze Age (3300–1800 BCE), Iron Age (1800–200 BCE), Classical Period (200–650 AD), Early Medieval Period (650–1200 AD), Late Medieval Period (1200–1526 AD), Early Modern Period (1526–1857 AD), and Modern India (1857 AD–present). To construct 600 multiple-choice questions for the TAB-VLM benchmark, we randomly sampled images from the corresponding historical periods.

\begin{table}[htbp]
\centering
\resizebox{0.5\textwidth}{!}{%
\begin{tabular}{lcc}
\toprule
\textbf{Historical Period} & \textbf{Time Range} & \textbf{Artifacts} \\
\midrule
Prehistoric Period & Before 3300 BCE & 276 \\
Bronze Age (Indus Valley Civilization) & 3300 -- 1800 BCE & 150 \\
Iron Age & 1800 -- 200 BCE & 259 \\
Classical Period & 200 -- 650 AD & 310 \\
Early Medieval Period & 650 -- 1200 AD & 139 \\
Late Medieval Period & 1200 -- 1526 AD & 155 \\
Early Modern Period & 1526 -- 1857 AD & 137 \\
Modern India & 1857 AD -- Present & 174 \\
\midrule
\textbf{Total} & \textbf{--} & \textbf{1,600} \\
\bottomrule
\end{tabular}%
}
\caption{Distribution of Artifacts Across Historical Periods in the TAB-VLM Benchmark}
\label{tab:period_distribution}
\end{table}

We implement a data cleaning process that included expert validation by the authors of this paper. This process involved eliminating duplicates, removing artifacts with ambiguous or disputed dating, and filtering out items lacking sufficient visual detail for reliable temporal assessment. As a result, the corpus was significantly reduced while maintaining strict quality standards for historical accuracy. The final cleaned dataset consists of 1,600 artifacts.

\begin{figure}[htbp]
    \centering
    \includegraphics[width=0.98\linewidth]{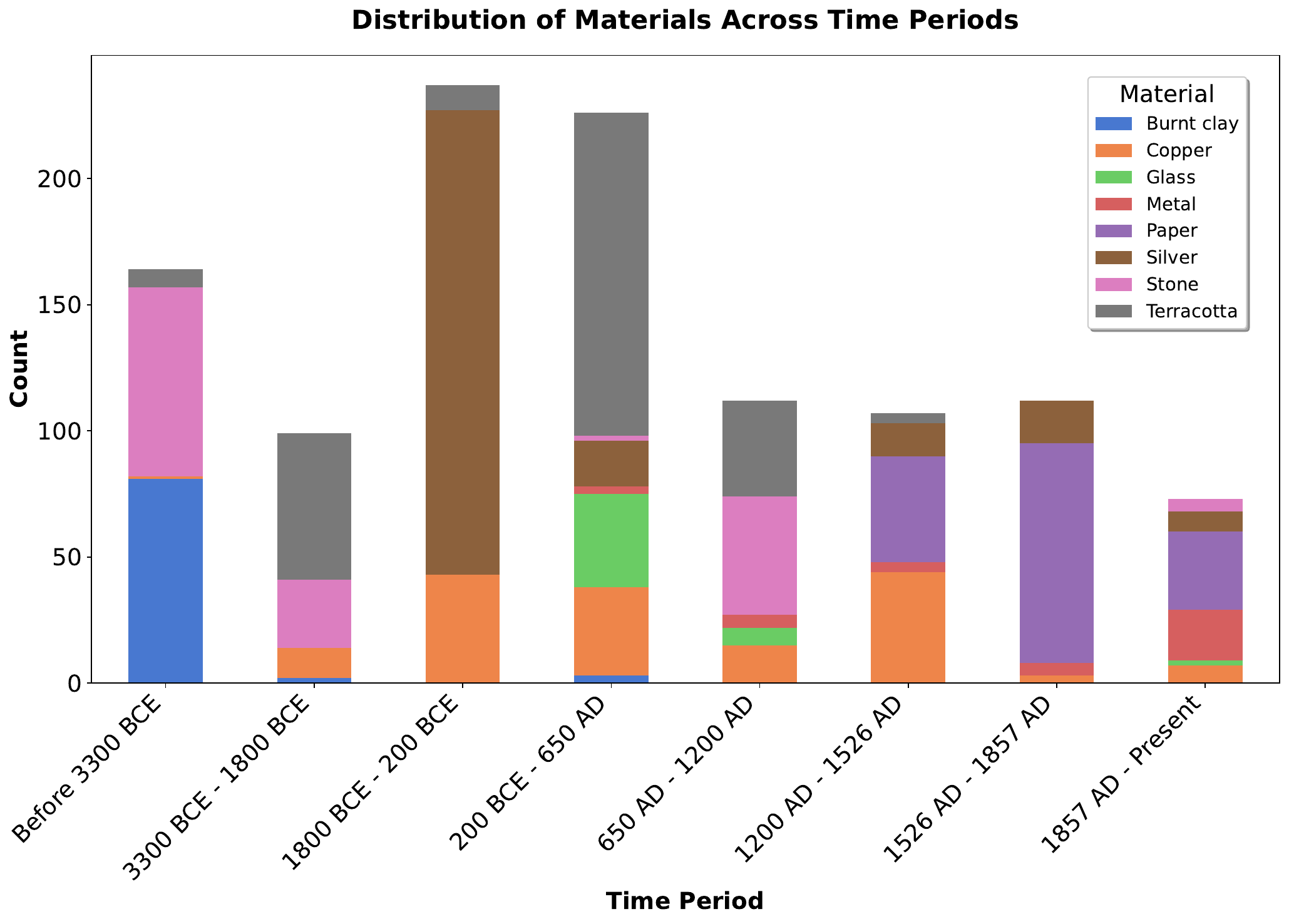}
    \caption{Distribution of materials across historical time periods showing predominance of stone artifacts in prehistoric periods and silver in medieval India.}
    \label{fig:materials_dist}
    \vspace{-0.15in}
\end{figure}

The final dataset exhibits a temporal distribution reflecting the archaeological record of Indian cultural heritage, with prehistoric periods containing 276 artifacts, representing the rich material culture of early Indian civilizations (Table \ref{tab:period_distribution}). Medieval periods are well-represented with 139-155 artifacts each, while more recent periods contain fewer items due to different preservation patterns and the scope of museum collections. The distribution of object types reveals clear temporal patterns: terracotta artifacts dominate prehistoric collections, reflecting early ceramic traditions; manuscripts become prominent during the Late Medieval Period, showcasing court literature and illuminated texts; and stone sculptures span multiple eras, particularly flourishing during the Classical Period (Figure \ref{fig:object_types}). Material composition demonstrates technological evolution across Indian history, with stone artifacts prevalent in prehistoric periods, copper and bronze emerging in ancient times, and paper becoming prominent in medieval and modern India (Figure \ref{fig:materials_dist}). This material diversity provides crucial ground truth for evaluating models' understanding of technological development timelines and material availability across different historical periods.

\section{Experimental Results}

\subsection{Experimental Setup}
We conduct our experiments using ten different VLMs, comprising six open-source and four proprietary models to provide comprehensive coverage of the current VLM landscape. The open-source models include instruction-tuned variants from the Qwen2-VL~\citep{wang2024qwen2vl} series with 2B and 7B parameters, the Qwen2.5-VL~\citep{bai2025qwen2.5vl} series with 3B and 7B parameters, and InternVL3~\citep{zhu2025internvl3} models with 2B and 8B parameters, representing state-of-the-art open-source capabilities across different model scales. As proprietary baselines, we evaluate OpenAI's GPT-5.2~\citep{openai2025gpt5.2}, GPT-5-mini~\citep{openai2025gpt5}, GPT-4o~\citep{hurst2024gpt4o} and GPT-4o-mini~\citep{hurst2024gpt4o}, which represent the current frontier in commercial VLM performance. All open-source model experiments are conducted on a single NVIDIA A100 GPU using the default hyperparameters from the HuggingFace~\citep{wolf2019huggingface} implementation with default hyperparameters. 
All models are evaluated using the same prompt templates (prompt provided in Appendix \ref{sec:prompt_template}) and evaluation pipeline to ensure fair comparison, with each question processed independently and model responses parsed using consistent answer extraction rules tailored to each question type.

\subsection{Evaluation Metric}
We evaluate model performance using accuracy as the primary metric, calculated as the proportion of correctly answered questions within each task category and overall. For single-choice questions (Style-Period Attribution, Odd-One-Out Period Detection), accuracy represents exact match between predicted and ground truth answers. For multi-choice questions (Manufacturing Technique, Material Availability), we require exact set matching where all correct options must be selected and no incorrect options chosen. For sequence-based tasks (Chronological Sequencing), accuracy measures exact ordering match, while for grouping tasks (Period Grouping), we require precise identification of all three artifacts from the same historical period. Standard deviation is computed across multiple evaluation runs to assess result stability.

\subsection{Results and Analysis}
Our experiment reveals several significant problems in temporal reasoning capabilities of VLMs when interpreting historical artifacts. Table \ref{tab:vlm_comparison_updated} presents accuracy results for each model across all task categories, demonstrating substantial performance gaps between proprietary and open-source models. GPT-5.2 achieved the highest overall accuracy at 58.7\%, establishing clear superiority across most task categories and representing the current state-of-the-art for temporal reasoning in cultural heritage contexts. The model demonstrated exceptional performance in Material Availability detection (92.1\%) and Style-Period Attribution (65.0\%), indicating robust understanding of technological constraints and stylistic recognition capabilities that enable identification of anachronistic materials and period-appropriate artistic elements. However, even this best-performing model exhibited limitations in temporal ordering tasks, achieving only 37.2\% accuracy in Chronological Sequencing and 45.2\% in Period Grouping, revealing gaps in understanding temporal relationships between artifacts across different historical periods. GPT-5-mini~\cite{openai2025gpt5} and GPT-4o-mini achieved competitive overall performance at 51.2\% and 47.2\% respectively. GPT-4o-mini notably surpasses GPT-4o in Material Availability (89.0\% vs 85.3\%) and Odd-One-Out Period Detection (48.0\% vs 45.3\%), suggesting that cost-effective models can achieve comparable or superior performance on specific temporal reasoning subtasks. Among open-source alternatives, Qwen2.5-VL-7B emerged as the strongest performer with 42.6\% overall accuracy, demonstrating particularly strong capabilities in Style-Period Attribution (58.8\%) and achieving the highest open-source performance in Odd-One-Out Period Detection (51.0\%), indicating effective training on temporal pattern recognition despite being substantially smaller than proprietary models. InternVL3-8B achieved 36.2\% overall accuracy with balanced performance across most categories, while Qwen2-VL-7B reached 32.8\% overall accuracy but showed pronounced weaknesses in Manufacturing Technique recognition (16.7\%) and Period Grouping (10.0\%), suggesting specific limitations in understanding technological evolution and temporal clustering. Smaller parameter models consistently underperformed, with Qwen2.5-VL-3B achieving 29.0\% accuracy and Qwen2-VL-2B reaching only 19.7\%, though both maintained reasonable Style-Period Attribution capabilities (52.5\% and 47.5\% respectively), indicating that certain temporal reasoning capabilities may be robust to model scale reduction. Across all evaluated models, Material Availability emerged as the most accessible task category, with six of ten models achieving above 45\% accuracy and both GPT variants exceeding 85\%, suggesting that models can effectively identify anachronistic material usage through learned associations between materials and historical periods. Conversely, Chronological Sequencing proved universally challenging, with no model exceeding 37.2\% accuracy and most open-source models performing below 20\%, indicating problems in understanding temporal progression and artifact dating relationships. Period Grouping similarly challenged all models, with performance ranging from 8.0\% to 45.2\%, revealing difficulties in recognizing shared temporal characteristics across different artifact types and cultural contexts. Notably, the relationship between model scale and temporal reasoning performance proved complex and non-linear, as evidenced by Qwen2.5-VL-3B achieving comparable or superior performance to larger models in several categories, while InternVL3-8B showed only marginal improvements over InternVL3-2B (36.2\% vs 30.5\%), suggesting that architectural innovations, training methodology, and dataset composition may be more influential factors than raw parameter count for developing effective temporal reasoning capabilities in VLMs.

\section{Discussion}
\vspace{-0.1in}
Our evaluation reveals fundamental limitations in current VLMs' ability to reason temporally about historical artifacts. We address our three research questions through quantitative results and qualitative error analysis, providing diagnostic insights into the nature of cultural anachronism in multimodal AI systems.

\vspace{-0.1in}
\subsection{Extent of Cultural Anachronism Across Model Architectures}

Addressing RQ1, current state-of-the-art VLMs exhibit substantial cultural anachronism when interpreting historical artifacts, with even the best-performing model (GPT-5.2) achieving only 58.7\% overall accuracy, a modest improvement over the random baseline of 12.8\%. This 45.9 percentage point gap, while substantial, reveals that even frontier models struggle significantly with temporal reasoning. The consistency of poor performance across different model architectures and scales suggests that cultural anachronism represents a fundamental limitation rather than a scaling problem. Notably, GPT-4o's marginal advantage over GPT-4o-mini (50.4\% vs 47.2\%), and Qwen2.5-VL-3B's competitive performance with much larger models in several categories, indicate that architectural innovations and training methodology may be more critical than raw parameter count. These findings contrast sharply with VLMs' generally strong performance on contemporary visual understanding benchmarks, suggesting that temporal reasoning about historical contexts requires fundamentally different capabilities than those emphasized in current training paradigms.

\subsection{Task-Specific Challenges and Error Patterns}

Addressing RQ2, we observe pronounced variations in performance across task categories that reveal distinct failure modes. 

\textbf{Chronological Sequencing} emerged as the most challenging task, with the best model achieving only 37.2\% accuracy compared to a random baseline of 4.2\%. Qualitative analysis reveals systematic errors: models frequently confuse adjacent historical periods (e.g., Classical vs. Early Medieval) and show a recency bias, often placing more weathered artifacts as "older" regardless of their actual period. For instance, GPT-5.2 incorrectly ordered a well-preserved Bronze Age Indus seal after a weathered Classical period sculpture, suggesting reliance on surface condition rather than stylistic or technological markers. \textbf{Material Availability} proved most accessible (GPT-5.2: 92.1\% vs. random baseline: 6.25\%), but analysis of the 7.9\% error rate reveals concerning patterns. Models occasionally attribute modern materials (synthetic dyes, industrial alloys) to medieval artifacts, or fail to recognize that certain precious metals became available through trade routes in specific periods. This suggests models may rely on general historical knowledge about material development rather than nuanced understanding of regional technological timelines.
\textbf{Style-Period Attribution} showed intermediate difficulty (GPT-5.2: 65.0\% vs. random baseline: 25.0\%). Error analysis reveals systematic confusion between adjacent periods, with 68\% of misclassifications occurring within ±1 period boundary. Models particularly struggle with transitional artifacts showing mixed stylistic elements, and exhibit a tendency to over-classify artifacts into later periods when visual details are ambiguous, a potential consequence of training data skew toward more recent, better-documented artifacts.
\textbf{Period Grouping} challenged all models (best: 45.2\%; random: 10.0\%), with failures revealing an inability to recognize coherent technological and stylistic signatures across different artifact types. Models often group visually similar artifacts (e.g., all terracotta items) regardless of period, suggesting that superficial feature matching overrides temporal reasoning. These patterns indicate that VLMs perform better on tasks solvable through memorized factual associations (material-to-period mappings) than those requiring genuine temporal cognition and visual pattern recognition across stylistic evolution.

\subsection{Implications for Development and Deployment}

Addressing RQ3, the disconnect between VLMs' strong general visual capabilities and poor temporal reasoning reveals fundamental gaps in current training paradigms. Visual feature extraction enabling modern object detection does not transfer effectively to historical contexts requiring period-appropriate interpretive frameworks.

\textbf{Risks for Cultural Heritage Applications and Directions for Mitigation.} VLMs deployed in museum digitization or educational platforms may systematically misrepresent artifacts, potentially perpetuating colonial-era anachronistic interpretations or erasing nuanced historical developments. The observed issues are particularly problematic for non-Western cultural heritage. Our findings also suggest several avenues for improvement: (1) \textit{Training data curation}: Explicitly include temporally annotated historical corpora with diverse cultural contexts and period coverage; (2) \textit{Specialized objectives}: Incorporate contrastive learning tasks that require distinguishing adjacent periods and recognizing anachronistic material-technique combinations; (3) \textit{Evaluation protocols}: Systematically assess cultural anachronism before deployment, particularly for applications involving non-Western heritage.

\vspace{-0.05in}
\section{Conclusion}
\vspace{-0.05in}
We introduce TAB-VLM, the first benchmark to systematically evaluate cultural anachronism in VLMs through 600 questions across 1,600 Indian cultural artifacts. Our evaluation of ten state-of-the-art VLMs reveals widespread temporal reasoning deficiencies, with even the best model (GPT-5.2) achieving only 58.7\% overall accuracy and overall struggle in chronological sequencing tasks ($\leq 38\%$ accuracy). The consistency of poor performance across different architectures and scales indicates that cultural anachronism represents a fundamental problem in current VLM training paradigms rather than a scaling problem. These findings have several important implications for cultural heritage applications, where anachronistic interpretations risk misrepresenting historical artifacts and perpetuating biased cultural narratives. Our work establishes cultural anachronism as an essential evaluation dimension for multimodal AI and provides a framework for developing temporally-aware systems capable of respectful engagement with cultural heritage materials. We hope that TAB-VLM will catalyze further research into incorporating temporal reasoning and cultural sensitivity into VLM training objectives, particularly for underrepresented non-Western heritage. 

\newpage
\section{Limitations and Future Work}\label{sec:limitations}
Our evaluation is constrained by several methodological limitations that affect generalizability. The benchmark focuses exclusively on Indian cultural artifacts, limiting cross-cultural applicability and potentially missing anachronistic patterns specific to other cultural contexts. Our visual-only evaluation approach excludes textual metadata that would typically accompany artifacts in real deployments. Expert annotations, though rigorously validated, reflect particular scholarly perspectives and discrete temporal classifications that may not accommodate transitional or culturally mixed artifacts. Additionally, our model selection represents only a subset of available VLMs, and the rapid pace of development means newer architectures may exhibit different anachronistic patterns than those evaluated here.

The modest performance differences across model scales indicate that simply scaling existing approaches is insufficient. Apart from improving the dataset discussed above, future work should also investigate model development techniques such as whether fine-tuning on temporally annotated historical datasets can reduce anachronism, and whether incorporating explicit temporal reasoning signals (e.g., stratified period embeddings) during training improves performance. Additionally, cross-cultural validation on artifacts from multiple cultural traditions would clarify whether the observed patterns generalize or reflect culture-specific training data gaps.

\section{Use of Language Models}
Large language models were used in a limited capacity to assist with minor editing and polishing of the manuscript. The use of LLMs was strictly limited to text formatting and grammatical corrections; no AI tools were used to generate scientific ideas, interpret results, or formulate the core arguments of this work. All technical content, experimental design, results, and conclusions were produced, verified, and finalized by the authors.


\bibliography{custom}

@misc{museumsofindia2025,
  author       = {{Sahapedia}},
  title        = {Museums of India},
  year         = {2025},
  url          = {https://www.museumsofindia.org/},
  note         = {Accessed: 2025-08-02},
  howpublished = {\url{https://www.museumsofindia.org/}},
  publisher    = {Sahapedia}
}

@article{liu2023llavavisual,
  title={Visual instruction tuning},
  author={Liu, Haotian and Li, Chunyuan and Wu, Qingyang and Lee, Yong Jae},
  journal={Advances in neural information processing systems},
  volume={36},
  pages={34892--34916},
  year={2023}
}

@inproceedings{liu2024improvedllava,
  title={Improved baselines with visual instruction tuning},
  author={Liu, Haotian and Li, Chunyuan and Li, Yuheng and Lee, Yong Jae},
  booktitle={Proceedings of the IEEE/CVF Conference on Computer Vision and Pattern Recognition},
  pages={26296--26306},
  year={2024}
}

@article{hurst2024gpt4o,
  title={Gpt-4o system card},
  author={Hurst, Aaron and Lerer, Adam and Goucher, Adam P and Perelman, Adam and Ramesh, Aditya and Clark, Aidan and Ostrow, AJ and Welihinda, Akila and Hayes, Alan and Radford, Alec and others},
  journal={arXiv preprint arXiv:2410.21276},
  year={2024}
}

@article{wang2024qwen2vl,
  title={Qwen2-vl: Enhancing vision-language model's perception of the world at any resolution},
  author={Wang, Peng and Bai, Shuai and Tan, Sinan and Wang, Shijie and Fan, Zhihao and Bai, Jinze and Chen, Keqin and Liu, Xuejing and Wang, Jialin and Ge, Wenbin and others},
  journal={arXiv preprint arXiv:2409.12191},
  year={2024}
}

@article{bai2025qwen2.5vl,
  title={Qwen2. 5-VL Technical Report},
  author={Bai, Shuai and Chen, Keqin and Liu, Xuejing and Wang, Jialin and Ge, Wenbin and Song, Sibo and Dang, Kai and Wang, Peng and Wang, Shijie and Tang, Jun and others},
  journal={arXiv preprint arXiv:2502.13923},
  year={2025}
}

@inproceedings{radford2021learning,
  title={Learning transferable visual models from natural language supervision},
  author={Radford, Alec and Kim, Jong Wook and Hallacy, Chris and Ramesh, Aditya and Goh, Gabriel and Agarwal, Sandhini and Sastry, Girish and Askell, Amanda and Mishkin, Pamela and Clark, Jack and others},
  booktitle={International conference on machine learning},
  pages={8748--8763},
  year={2021},
  organization={PmLR}
}

@article{yu2022coca,
  title={Coca: Contrastive captioners are image-text foundation models},
  author={Yu, Jiahui and Wang, Zirui and Vasudevan, Vijay and Yeung, Legg and Seyedhosseini, Mojtaba and Wu, Yonghui},
  journal={arXiv preprint arXiv:2205.01917},
  year={2022}
}

@article{li2024culturellm,
  title={Culturellm: Incorporating cultural differences into large language models},
  author={Li, Cheng and Chen, Mengzhuo and Wang, Jindong and Sitaram, Sunayana and Xie, Xing},
  journal={Advances in Neural Information Processing Systems},
  volume={37},
  pages={84799--84838},
  year={2024}
}

@article{hwang2025cats,
  title={CATS: cultural-heritage classification using LLMs and distribute model},
  author={Hwang, Hyerin and Park, Chan-Woo and Kim, Hee-Kwon and Lee, Jae-Ho},
  journal={npj Heritage Science},
  volume={13},
  number={1},
  pages={76},
  year={2025},
  publisher={Springer International Publishing Cham}
}

@inproceedings{trichopoulos2023large,
  title={Large language models for cultural heritage},
  author={Trichopoulos, Georgios},
  booktitle={Proceedings of the 2nd International Conference of the ACM Greek SIGCHI Chapter},
  pages={1--5},
  year={2023}
}

@article{arnold2024explainable,
  title={Explainable Search and Discovery of Visual Cultural Heritage Collections with Multimodal Large Language Models},
  author={Arnold, Taylor and Tilton, Lauren},
  journal={arXiv preprint arXiv:2411.04663},
  year={2024}
}

@article{ghaboura2025time,
  title={Time Travel: A Comprehensive Benchmark to Evaluate LMMs on Historical and Cultural Artifacts},
  author={Ghaboura, Sara and More, Ketan and Thawkar, Ritesh and Alghallabi, Wafa and Thawakar, Omkar and Khan, Fahad Shahbaz and Cholakkal, Hisham and Khan, Salman and Anwer, Rao Muhammad},
  journal={arXiv preprint arXiv:2502.14865},
  year={2025}
}

@article{liang2022holistic,
  title={Holistic evaluation of language models},
  author={Liang, Percy and Bommasani, Rishi and Lee, Tony and Tsipras, Dimitris and Soylu, Dilara and Yasunaga, Michihiro and Zhang, Yian and Narayanan, Deepak and Wu, Yuhuai and Kumar, Ananya and others},
  journal={arXiv preprint arXiv:2211.09110},
  year={2022}
}

@article{bhatia2024datelogicqa,
  title={DateLogicQA: Benchmarking Temporal Biases in Large Language Models},
  author={Bhatia, Gagan and Tang, MingZe and Mahanta, Cristina and Kazi, Madiha},
  journal={arXiv preprint arXiv:2412.13377},
  year={2024}
}

@article{gallegos2024bias,
  title={Bias and fairness in large language models: A survey},
  author={Gallegos, Isabel O and Rossi, Ryan A and Barrow, Joe and Tanjim, Md Mehrab and Kim, Sungchul and Dernoncourt, Franck and Yu, Tong and Zhang, Ruiyi and Ahmed, Nesreen K},
  journal={Computational Linguistics},
  volume={50},
  number={3},
  pages={1097--1179},
  year={2024},
  publisher={MIT Press 255 Main Street, 9th Floor, Cambridge, Massachusetts 02142, USA~…}
}

@article{ko2023large,
  title={Large language models are temporal and causal reasoners for video question answering},
  author={Ko, Dohwan and Lee, Ji Soo and Kang, Wooyoung and Roh, Byungseok and Kim, Hyunwoo J},
  journal={arXiv preprint arXiv:2310.15747},
  year={2023}
}

@article{siliutina2024cultural,
  title={Cultural preservation and digital heritage: challenges and opportunities},
  author={Siliutina, Iryna and Tytar, Olena and Barbash, Marina and Petrenko, Nataliia and Yepyk, Larysa},
  journal={Amazonia Investiga},
  volume={13},
  number={75},
  pages={262--273},
  year={2024}
}

@article{birhane2021multimodal,
  title={Multimodal datasets: misogyny, pornography, and malignant stereotypes},
  author={Birhane, Abeba and Prabhu, Vinay Uday and Kahembwe, Emmanuel},
  journal={arXiv preprint arXiv:2110.01963},
  year={2021}
}

@article{thylstrup2022ethics,
  title={The ethics and politics of data sets in the age of machine learning: Deleting traces and encountering remains},
  author={Thylstrup, Nanna Bonde},
  journal={Media, Culture \& Society},
  volume={44},
  number={4},
  pages={655--671},
  year={2022},
  publisher={SAGE Publications Sage UK: London, England}
}

@article{li2025benchmark,
  title={A Survey of State of the Art Large Vision Language Models: Alignment, Benchmark, Evaluations and Challenges},
  author={Li, Zongxia and Wu, Xiyang and Du, Hongyang and Liu, Fuxiao and Nghiem, Huy and Shi, Guangyao},
  year={2025},
  journal={arXiv preprint arXiv:2501.02189},
}

@article{chang2024survey,
  title={A survey on evaluation of large language models},
  author={Chang, Yupeng and Wang, Xu and Wang, Jindong and Wu, Yuan and Yang, Linyi and Zhu, Kaijie and Chen, Hao and Yi, Xiaoyuan and Wang, Cunxiang and Wang, Yidong and others},
  journal={ACM transactions on intelligent systems and technology},
  volume={15},
  number={3},
  pages={1--45},
  year={2024},
  publisher={ACM New York, NY}
}

@inproceedings{sambasivan2021re,
  title={Re-imagining algorithmic fairness in india and beyond},
  author={Sambasivan, Nithya and Arnesen, Erin and Hutchinson, Ben and Doshi, Tulsee and Prabhakaran, Vinodkumar},
  booktitle={Proceedings of the 2021 ACM conference on fairness, accountability, and transparency},
  pages={315--328},
  year={2021}
}

@article{kannen2024beyond,
  title={Beyond aesthetics: Cultural competence in text-to-image models},
  author={Kannen, Nithish and Ahmad, Arif and Andreetto, Marco and Prabhakaran, Vinodkumar and Prabhu, Utsav and Dieng, Adji Bousso and Bhattacharyya, Pushpak and Dave, Shachi},
  journal={arXiv preprint arXiv:2407.06863},
  year={2024}
}

@article{wang2023tram,
  title={Tram: Benchmarking temporal reasoning for large language models},
  author={Wang, Yuqing and Zhao, Yun},
  journal={arXiv preprint arXiv:2310.00835},
  year={2023}
}

@inproceedings{jain2023language,
  title={Do language models have a common sense regarding time? revisiting temporal commonsense reasoning in the era of large language models},
  author={Jain, Raghav and Sojitra, Daivik and Acharya, Arkadeep and Saha, Sriparna and Jatowt, Adam and Dandapat, Sandipan},
  booktitle={Proceedings of the 2023 Conference on Empirical Methods in Natural Language Processing},
  pages={6750--6774},
  year={2023}
}

@article{rao2024normad,
  title={Normad: A benchmark for measuring the cultural adaptability of large language models},
  author={Rao, Abhinav and Yerukola, Akhila and Shah, Vishwa and Reinecke, Katharina and Sap, Maarten},
  journal={arXiv preprint arXiv:2404.12464},
  year={2024}
}

@article{fung2024massively,
  title={Massively multi-cultural knowledge acquisition \& lm benchmarking},
  author={Fung, Yi and Zhao, Ruining and Doo, Jae and Sun, Chenkai and Ji, Heng},
  journal={arXiv preprint arXiv:2402.09369},
  year={2024}
}

@article{kharchenko2024well,
  title={How well do llms represent values across cultures? empirical analysis of llm responses based on hofstede cultural dimensions},
  author={Kharchenko, Julia and Roosta, Tanya and Chadha, Aman and Shah, Chirag},
  journal={arXiv preprint arXiv:2406.14805},
  year={2024}
}

@article{chiu2024culturalbench,
  title={Culturalbench: a robust, diverse and challenging benchmark on measuring the (lack of) cultural knowledge of llms},
  author={Chiu, Yu Ying and Jiang, Liwei and Lin, Bill Yuchen and Park, Chan Young and Li, Shuyue Stella and Ravi, Sahithya and Bhatia, Mehar and Antoniak, Maria and Tsvetkov, Yulia and Shwartz, Vered and others},
  journal={arXiv preprint arXiv:2410.02677},
  year={2024}
}

@article{li2024culturepark,
  title={CulturePark: Boosting Cross-cultural Understanding in Large Language Models},
  author={Li, Cheng and Teney, Damien and Yang, Linyi and Wen, Qingsong and Xie, Xing and Wang, Jindong},
  journal={arXiv preprint arXiv:2405.15145},
  year={2024}
}

@article{adilazuarda2024towards,
  title={Towards measuring and modeling" culture" in llms: A survey},
  author={Adilazuarda, Muhammad Farid and Mukherjee, Sagnik and Lavania, Pradhyumna and Singh, Siddhant and Aji, Alham Fikri and O'Neill, Jacki and Modi, Ashutosh and Choudhury, Monojit},
  journal={arXiv preprint arXiv:2403.15412},
  year={2024}
}

@article{liu2023multilingual,
  title={Are multilingual llms culturally-diverse reasoners? an investigation into multicultural proverbs and sayings},
  author={Liu, Chen Cecilia and Koto, Fajri and Baldwin, Timothy and Gurevych, Iryna},
  journal={arXiv preprint arXiv:2309.08591},
  year={2023}
}

@article{hwang2023aligning,
  title={Aligning language models to user opinions},
  author={Hwang, EunJeong and Majumder, Bodhisattwa Prasad and Tandon, Niket},
  journal={arXiv preprint arXiv:2305.14929},
  year={2023}
}

@article{su2024timo,
  title={Timo: Towards Better Temporal Reasoning for Language Models},
  author={Su, Zhaochen and Zhang, Jun and Zhu, Tong and Qu, Xiaoye and Li, Juntao and Zhang, Min and Cheng, Yu},
  journal={arXiv preprint arXiv:2406.14192},
  year={2024}
}

@article{fatemi2024test,
  title={Test of Time: A Benchmark for Evaluating LLMs on Temporal Reasoning},
  author={Fatemi, Bahare and Kazemi, Mehran and Tsitsulin, Anton and Malkan, Karishma and Yim, Jinyeong and Palowitch, John and Seo, Sungyong and Halcrow, Jonathan and Perozzi, Bryan},
  journal={arXiv preprint arXiv:2406.09170},
  year={2024}
}

@inproceedings{yuan2024back,
  title={Back to the future: Towards explainable temporal reasoning with large language models},
  author={Yuan, Chenhan and Xie, Qianqian and Huang, Jimin and Ananiadou, Sophia},
  booktitle={Proceedings of the ACM on Web Conference 2024},
  pages={1963--1974},
  year={2024}
}

@article{nayak2024benchmarking,
  title={Benchmarking vision language models for cultural understanding},
  author={Nayak, Shravan and Jain, Kanishk and Awal, Rabiul and Reddy, Siva and van Steenkiste, Sjoerd and Hendricks, Lisa Anne and Sta{\'n}czak, Karolina and Agrawal, Aishwarya},
  journal={arXiv preprint arXiv:2407.10920},
  year={2024}
}

@article{liu2025culturevlm,
  title={CultureVLM: Characterizing and Improving Cultural Understanding of Vision-Language Models for over 100 Countries},
  author={Liu, Shudong and Jin, Yiqiao and Li, Cheng and Wong, Derek F and Wen, Qingsong and Sun, Lichao and Chen, Haipeng and Xie, Xing and Wang, Jindong},
  journal={arXiv preprint arXiv:2501.01282},
  year={2025}
}

@article{underwood2025can,
  title={Can Language Models Represent the Past without Anachronism?},
  author={Underwood, Ted and Nelson, Laura K and Wilkens, Matthew},
  journal={arXiv preprint arXiv:2505.00030},
  year={2025}
}

@article{chu2023timebench,
  title={Timebench: A comprehensive evaluation of temporal reasoning abilities in large language models},
  author={Chu, Zheng and Chen, Jingchang and Chen, Qianglong and Yu, Weijiang and Wang, Haotian and Liu, Ming and Qin, Bing},
  journal={arXiv preprint arXiv:2311.17667},
  year={2023}
}

@article{wang2025timecausality,
  title={TimeCausality: Evaluating the Causal Ability in Time Dimension for Vision Language Models},
  author={Wang, Zeqing and Zhang, Shiyuan and Tang, Chengpei and Wang, Keze},
  journal={arXiv preprint arXiv:2505.15435},
  year={2025}
}

@inproceedings{chen2024rextime,
  title={Rextime: A benchmark suite for reasoning-across-time in videos},
  author={Chen, Jr-Jen and Liao, Yu-Chien and Lin, Hsi-Che and Yu, Yu-Chu and Chen, Yen-Chun and Wang, Frank},
  booktitle={Advances in Neural Information Processing Systems},
  volume={37},
  pages={28662--28673},
  year={2024}
}

@article{padlewski2024vibeeval,
  title={Vibe-eval: A hard evaluation suite for measuring progress of multimodal language models},
  author={Padlewski, Piotr and Bain, Max and Henderson, Matthew and Zhu, Zhongkai and Relan, Nishant and Pham, Hai and Ong, Donovan and others},
  journal={arXiv preprint arXiv:2405.02287},
  year={2024}
}

@article{wolf2019huggingface,
  title={HuggingFace's Transformers: State-of-the-art Natural Language Processing},
  author={Wolf, Thomas and Debut, Lysandre and Sanh, Victor and Chaumond, Julien and Delangue, Cl{\'e}ment and Moi, Anthony and Cistac, Pierric and Rault, Tim and Louf, R{\'e}mi and Funtowicz, Morgan and Brew, Joe},
  journal={arXiv preprint arXiv:1910.03771},
  year={2019}
}

@article{zhu2025internvl3,
  title={InternVL3: Exploring Advanced Training and Test-Time Recipes for Open-Source Multimodal Models},
  author={Zhu, Jinguo and Wang, Weiyun and Chen, Zhe and Liu, Zhaoyang and Ye, Shenglong and Gu, Lixin and Tian, Hao and others},
  journal={arXiv preprint arXiv:2504.10479},
  year={2025}
}

@book{mcleod2015history,
  title={The History of India},
  author={McLeod, John},
  year={2015},
  publisher={Bloomsbury Publishing USA}
}

@misc{openai2025gpt5,
  author       = {OpenAI},
  title        = {GPT-5 System Card},
  year         = {2025},
  url          = {https://openai.com/index/gpt-5-system-card/},
  note         = {Published Aug~13,~2025; system card for the GPT-5 model family},
}

@misc{openai2025gpt5.2,
  author       = {OpenAI},
  title        = {Update to GPT-5 System Card: GPT-5.2},
  year         = {2025},
  url          = {https://openai.com/index/gpt-5-system-card-update-gpt-5-2/},
  note         = {Published Dec~11,~2025; update describing GPT-5.2 model family},
}

@article{upadhyay2025time,
  title={Time Blindness: Why Video-Language Models Can't See What Humans Can?},
  author={Upadhyay, Ujjwal and Ranjan, Mukul and Shen, Zhiqiang and Elhoseiny, Mohamed},
  journal={arXiv preprint arXiv:2505.24867},
  year={2025}
}

@article{imam2025can,
  title={Can Multimodal LLMs do Visual Temporal Understanding and Reasoning? The answer is No!},
  author={Imam, Mohamed Fazli and Lyu, Chenyang and Aji, Alham Fikri},
  journal={arXiv preprint arXiv:2501.10674},
  year={2025}
}

@article{landis1977measurement,
  title={The measurement of observer agreement for categorical data},
  author={Landis, J Richard and Koch, Gary G},
  journal={biometrics},
  pages={159--174},
  year={1977},
  publisher={JSTOR}
}

\clearpage
\newpage

\appendix

\section*{Appendix}
\label{sec:appendix}

\section{Task Specifications}
\label{sec:task_specs}

600 multiple-choice questions in our benchmark is equally distributed across six temporal reasoning task categories (100 questions per category). Each task type evaluates different aspects of temporal awareness and cultural anachronism detection. Below we provide complete specifications for each task, including the number of images, answer structure, random baseline performance, and distractor construction methodology.

\subsection{Chronological Sequencing}

This task evaluates fundamental temporal ordering capabilities by requiring models to arrange four artifacts from different historical periods in correct chronological sequence from oldest to newest. Models must recognize visual cues indicating technological sophistication, artistic style evolution, and material availability across time periods.

\textbf{Structure:} 4 artifact images presented simultaneously. \textbf{Answer format:} One correct ordering from 24 possible permutations (4! = 24). \textbf{Random baseline:} 4.2\% (1/24). \textbf{Distractor construction:} Artifacts are sampled from non-adjacent historical periods (minimum 2 periods apart) to ensure meaningful temporal separation and prevent trivial ordering based on obvious style differences.

\subsection{Odd-One-Out Period Detection}

This task assesses sensitivity to temporal anomalies by presenting four artifacts where three belong to one historical period and one belongs to a different period. Models must identify which artifact is temporally inconsistent with the others based on subtle visual cues in style, materials, manufacturing techniques, and artistic conventions.

\textbf{Structure:} 4 artifact images. \textbf{Answer format:} Single choice identifying the temporal outlier (1 of 4 options). \textbf{Random baseline:} 25.0\% (1/4). \textbf{Distractor construction:} Three artifacts from the same historical period plus one artifact from a temporally adjacent period to increase difficulty and avoid obvious mismatches.

\subsection{Period Grouping}

This task examines pattern recognition across stylistic and material elements by requiring identification of three contemporaneous artifacts from a set of five. Models must recognize coherent aesthetic traditions, technological signatures, and cultural expressions characteristic of specific historical periods.

\textbf{Structure:} 5 artifact images. \textbf{Answer format:} Select 3 artifacts from the same period (10 possible combinations: $\binom{5}{3}$ = 10). \textbf{Random baseline:} 10.0\% (1/10). \textbf{Distractor construction:} Three artifacts from the target period plus two artifacts from different adjacent periods, requiring careful visual analysis to distinguish shared temporal characteristics.

\subsection{Manufacturing Technique}

This task tests understanding of historical production methods by asking which manufacturing techniques would have been available when a given artifact was created. Models must demonstrate knowledge of technological development timelines and craftsmanship evolution across Indian cultural history.

\textbf{Structure:} 1 artifact image with 4 technique options. \textbf{Answer format:} Multi-select (select all applicable techniques). \textbf{Scoring:} Exact-match required, all correct options must be selected and no incorrect options chosen. \textbf{Random baseline:} 6.25\% (1/16 - one correct subset out of 16 possible selections).
\textbf{Distractor construction:} Options include period-appropriate techniques, anachronistic modern techniques, and techniques from adjacent historical periods.

\subsection{Material Availability}

This task evaluates knowledge of historical material science by identifying which materials would NOT have been available during an artifact's creation period. Models must understand material discovery timelines, trade route development, and technological constraints across different eras.

\textbf{Structure:} 1 artifact image with 4 material options. \textbf{Answer format:} Multi-select identifying anachronistic materials (select all that would NOT have been available). \textbf{Scoring:} Exact-match required. \textbf{Random baseline:} 6.25\% (1/16 - one correct subset out of 16 possible selections).
 \textbf{Distractor construction:} Options include period-appropriate materials, clearly anachronistic modern materials (e.g., plastic, acrylic for ancient artifacts), and materials from adjacent periods requiring nuanced knowledge of material availability timelines.

\subsection{Style-Period Attribution}

This task provides direct assessment of temporal classification capabilities by requiring models to match artifacts with their correct historical periods. Models must recognize distinctive visual characteristics, artistic conventions, and cultural expressions associated with specific temporal periods in Indian cultural history.

\textbf{Structure:} 1 artifact image with 4 period options. \textbf{Answer format:} Single choice from 4 historical period options. \textbf{Random baseline:} 25.0\% (1/4). \textbf{Distractor construction:} Four different historical periods spanning Indian history (e.g., Prehistoric, Modern India, Iron Age, Bronze Age), requiring fine-grained temporal discrimination.

\begin{table*}[t]
\centering
\small
\begin{tabular}{p{3.5cm}p{1.2cm}p{2.5cm}p{1.5cm}}
\toprule
\textbf{Task} & \textbf{Images} & \textbf{Answer Format} & \textbf{Random Baseline} \\
\midrule
Chronological Sequence & 4 & Ordered sequence (24 permutations) & 4.2\% \\
Odd-One-Out Period & 4 & Single choice (1 of 4) & 25.0\% \\
Period Grouping & 5 & Select 3 of 5 (10 combinations) & 10.0\% \\
Manufacturing Technique & 1 & Multi-select from 4 options & 6.25\% \\
Material Availability & 1 & Multi-select from 4 options & 6.25\% \\
Style-Period Attribution & 1 & Single choice (1 of 4 periods) & 25.0\% \\
\bottomrule
\end{tabular}
\caption{Summary of task specifications. Chance baselines computed for single-answer and sequence tasks; multi-select tasks use exact-match scoring where all correct options must be selected.}
\end{table*}

\textbf{Scoring methodology:} All tasks employ exact-match accuracy as the primary metric. For single-choice tasks (Odd-One-Out, Style-Period Attribution), the model must select the single correct option. For multi-select tasks (Manufacturing Technique, Material Availability), the model must select ALL correct options and NO incorrect options, partial credit is not awarded. For Chronological Sequencing, the model must produce the exact correct ordering. For Period Grouping, the model must identify precisely the three contemporaneous artifacts.

\section{Prompt Templates}
\label{sec:prompt_template}

This section presents the standardized prompt templates used for evaluating vision-language models on temporal reasoning tasks with cultural heritage artifacts. Each template is designed to elicit specific aspects of historical understanding while maintaining consistency across all evaluated models. The prompts use \texttt{<image>} tags to indicate visual inputs, with the actual images provided during inference.

\subsection{Chronological Sequence}

The chronological sequencing task evaluates a model's fundamental ability to perceive and understand temporal progression through visual features alone. Models receive four artifacts from distinct historical periods and must arrange them from oldest to newest. This requires recognizing evolutionary patterns in artistic styles, technological sophistication, material usage, and iconographic conventions. The task is particularly challenging because it demands not just period identification, but understanding the relative temporal relationships between multiple artifacts simultaneously. Success requires the model to construct a coherent timeline based on visual evidence of technological development, stylistic maturation, and cultural evolution across Indian history.

\begin{tcolorbox}[colback=chronologicallight!50, colframe=chronological, title=\textcolor{white}{Chronological Sequence Prompt Template}, coltitle=white, colbacktitle=chronological]
Below are 4 historical artifacts from different time periods.

Question: \{question\_text\}

Please arrange these artifacts in chronological order from oldest to newest:

A. \texttt{<image>}

B. \texttt{<image>}

C. \texttt{<image>}

D. \texttt{<image>}

Provide the sequence from oldest to newest using letters (e.g., "A, C, B, D").
\end{tcolorbox}

\subsection{Odd One Out Period}

This task assesses temporal anomaly detection by presenting four artifacts where three belong to the same historical period and one is from a different era. The challenge lies in identifying subtle visual cues that distinguish the outlier, whether through anachronistic stylistic elements, technological features, or material properties. Unlike simple classification tasks, this requires comparative analysis across multiple artifacts and recognition of shared temporal characteristics among the majority group. The task specifically tests whether models can detect deviations from expected period-consistent patterns, simulating real-world scenarios where cultural heritage experts must identify misattributed or misdated artifacts in museum collections.

\begin{tcolorbox}[colback=oddoneoutlight!50, colframe=oddoneout, title=\textcolor{white}{Odd One Out Period Prompt Template}, coltitle=white, colbacktitle=oddoneout]
Below are 4 historical artifacts. One of them belongs to a different time period than the others.

Question: \{question\_text\}

Which artifact belongs to a different time period than the others?

A. \texttt{<image>}

B. \texttt{<image>}

C. \texttt{<image>}

D. \texttt{<image>}

Answer with the option's letter (A, B, C, or D) from the given choices directly.
\end{tcolorbox}

\subsection{Material Availability}

The material availability task evaluates understanding of technological timelines and resource availability across historical periods. Given a single artifact image, models must identify which materials from a provided list would NOT have been available during the artifact's creation period. This requires knowledge of when specific materials were discovered, developed, or became accessible in the Indian subcontinent. For instance, distinguishing whether an artifact could have been made with bronze (available from 3300 BCE) versus steel (developed much later) requires understanding metallurgical history. The task directly tests for material-based anachronisms, a common form of temporal misattribution where modern materials are incorrectly associated with ancient artifacts. The multi-select format allows for multiple correct answers, as several materials may be anachronistic for any given period.

\begin{tcolorbox}[colback=materiallight!50, colframe=material, title=\textcolor{white}{Material Availability Prompt Template}, coltitle=white, colbacktitle=material]
Look at this historical artifact and answer the following question:

\texttt{<image>}

Question: \{question\_text\}

Please select all options that apply from the following choices:
\{options\_text\}

List all correct letters separated by commas (e.g., "A, C" if options A and C are correct).
\end{tcolorbox}

\subsection{Manufacturing Technique}

This task examines knowledge of historical production methods and technological capabilities. Models must determine which manufacturing techniques from a provided list would have been available when the artifact was created. The task requires understanding the evolution of craftsmanship methods, from primitive hand-molding and stone carving in prehistoric periods to sophisticated casting, glazing, and metalworking techniques in later eras. Success demands recognition of visual evidence indicating production methods, such as tool marks, construction joins, surface treatments, and structural features. The multi-select format reflects the reality that multiple techniques were often available simultaneously in different regions or for different purposes, requiring nuanced understanding of technological coexistence rather than simple chronological progression.

\begin{tcolorbox}[colback=manufacturinglight!50, colframe=manufacturing, title=\textcolor{white}{Manufacturing Technique Prompt Template}, coltitle=white, colbacktitle=manufacturing]
Look at this historical artifact and answer the following question:

\texttt{<image>}

Question: \{question\_text\}

Please select all options that apply from the following choices:
\{options\_text\}

List all correct letters separated by commas (e.g., "A, C" if options A and C are correct).
\end{tcolorbox}

\subsection{Period Grouping}

Period grouping tests pattern recognition across stylistic and material elements by requiring identification of three contemporaneous artifacts from a set of five. This task demands synthesis of multiple temporal cues, artistic conventions, technological signatures, material properties, and cultural motifs, to recognize coherent aesthetic traditions. Unlike the odd-one-out task which focuses on identifying a single outlier, this requires positive identification of shared temporal characteristics among multiple artifacts while recognizing that two artifacts belong to different periods. The task simulates museum curation scenarios where artifacts must be grouped for period-specific exhibitions, requiring both individual dating expertise and comparative temporal analysis.

\begin{tcolorbox}[colback=groupinglight!50, colframe=grouping, title=\textcolor{white}{Period Grouping Prompt Template}, coltitle=white, colbacktitle=grouping]
Below are 5 historical artifacts. Three of them were likely created during the same time period.

Question: \{question\_text\}

Which three artifacts were likely created during the same time period?

A. \texttt{<image>}

B. \texttt{<image>}

C. \texttt{<image>}

D. \texttt{<image>}

E. \texttt{<image>}

List the three letters corresponding to artifacts from the same period (e.g., "A, C, E").
\end{tcolorbox}

\subsection{Style-Period Attribution}

Style-period attribution represents the most direct test of temporal classification capabilities. Given a single artifact, models must assign it to the correct historical period from four options spanning different eras of Indian history. This task requires recognition of distinctive visual characteristics, artistic conventions, iconographic traditions, and material signatures associated with specific temporal periods. The four-option format forces discrimination between periods that may share some overlapping features, requiring models to identify the most diagnostic temporal markers. Success on this task indicates mastery of period-specific aesthetic vocabularies and understanding of how artistic expressions evolved across different phases of Indian cultural development.

\begin{tcolorbox}[colback=attributionlight!50, colframe=attribution, title=\textcolor{black}{Style Period Attribution Prompt Template}, coltitle=black, colbacktitle=attribution]
Look at this historical artifact and answer the following question:

\texttt{<image>}

Question: \{question\_text\}

Please select the correct option from the following choices:
\{options\_text\}

Answer with the option's letter (A, B, C, or D) from the given choices directly.
\end{tcolorbox}

\section{Additional Analyses}
\label{sec:additional_analyses}

This section provides supplementary analyses that complement the main results of Section~4. We examine (i) the factual-to-relational task spectrum, (ii) a manual failure-mode taxonomy over 96 incorrect GPT-4o responses, (iii) partial-credit evaluation, (iv) the reliability of our automatic evaluation pipeline, (v) the effects of temporal imbalance and textual metadata, and (vi) a cross-cultural pilot study on Western artifacts.

\subsection{Factual vs.\ Relational Task Spectrum}
\label{sec:spectrum}

The six tasks in TAB-VLM span a continuum from factual association (e.g., associating materials with historical periods) to relational temporal reasoning (e.g., ordering artifacts across periods). To quantify this continuum, we group the tasks into two categories: \emph{fact-dominant} tasks (Material Availability, Style-Period Attribution, Manufacturing Technique), which can in principle be solved via memorized material--era mappings; and \emph{temporal-dominant} tasks (Odd-One-Out Period, Period Grouping, Chronological Sequencing), which require reasoning over temporal relationships between multiple artifacts.

\begin{table}[h!]
\centering
\small
\resizebox{\linewidth}{!}{
\begin{tabular}{lccc}
\toprule
\textbf{Task} & \textbf{Type} & \textbf{GPT-4o} & \textbf{Qwen2.5-VL-7B} \\
\midrule
Material Availability & Fact-Dom. & 85.3 & 69.0 \\
Style-Period Attribution & Fact-Dom. & 60.3 & 58.8 \\
Manufacturing Technique & Fact-Dom. & 49.8 & 41.5 \\
Odd-One-Out Period & Temp-Dom. & 45.3 & 51.0 \\
Period Grouping & Temp-Dom. & 32.3 & 22.0 \\
Chronological Sequencing & Temp-Dom. & 30.7 & 14.0 \\
\bottomrule
\end{tabular}
}
\caption{Accuracy (\%) across the factual-to-relational task spectrum. Random baselines: Material Availability and Manufacturing Technique 6.25\%; Style-Period Attribution and Odd-One-Out Period 25.0\%; Period Grouping 10.0\%; Chronological Sequencing 4.2\%.}
\label{tab:task_spectrum}
\end{table}

Performance drops sharply from fact-dominant to temporal-dominant tasks. GPT-4o exhibits a 54.6-point gap between Material Availability (85.3\%) and Chronological Sequencing (30.7\%), and this gradient is consistent across all ten models evaluated in Table~1. The gap is not attributable to missing cultural knowledge alone: smaller open-source models such as Qwen2.5-VL-3B score only 25.4\% on Material Availability, indicating that the fact-dominant end is non-trivial at smaller scales. Conversely, even frontier models fail on temporal-dominant tasks (GPT-5.2: 37.2\% on Chronological Sequencing). This pattern suggests that \emph{relational temporal reasoning}, rather than \emph{cultural knowledge retrieval}, is the primary bottleneck in current VLMs on TAB-VLM.

\subsection{Failure Mode Taxonomy}
\label{sec:failure_modes}

To characterise the nature of VLM errors on TAB-VLM, we manually analysed 96 incorrect GPT-4o responses, sampled uniformly at 16 per task. Each response was classified into one of three mutually exclusive failure types:
\begin{itemize}
    \item \textbf{Knowledge gap}: incorrect or missing historical/cultural knowledge (e.g., wrong period attribution, incorrect material--era association).
    \item \textbf{Visual grounding failure}: misreading visual cues in the image (e.g., misidentifying stylistic features or material texture).
    \item \textbf{Relational temporal reasoning failure}: recognising individual artifacts correctly but failing at cross-artifact temporal comparison (e.g., describing each artifact's period correctly yet ordering them incorrectly).
\end{itemize}

\begin{table}[h!]
\centering
\small
\begin{tabular}{lc}
\toprule
\textbf{Failure Type} & \textbf{Frequency} \\
\midrule
Knowledge Gap & 29\% (28/96) \\
Visual Grounding Failure & 8\% (8/96) \\
Relational Temporal Reasoning Failure & 63\% (60/96) \\
\bottomrule
\end{tabular}
\caption{Overall failure-type distribution across 96 incorrect GPT-4o responses.}
\label{tab:failure_overall}
\end{table}

Relational temporal reasoning failures dominate at 63\%, followed by knowledge gaps at 29\% and a small minority of visual grounding failures at 8\%. The per-task breakdown (Table~\ref{tab:failure_pertask}) shows that failure types are tightly aligned with task category.

\begin{table}[h!]
\centering
\small
\resizebox{\columnwidth}{!}{%
\begin{tabular}{llccc}
\toprule
\textbf{Task} & \textbf{Category} & \textbf{Knowl.} & \textbf{Visual} & \textbf{Relational} \\
\midrule
Manufacturing Technique & Fact-Dom. & 100\% & 0\% & 0\% \\
Material Availability & Fact-Dom. & 88\% & 12\% & 0\% \\
Style-Period Attribution & Fact-Dom. & 75\% & 19\% & 6\% \\
Odd-One-Out Period & Temp-Dom. & 25\% & 38\% & 37\% \\
Period Grouping & Temp-Dom. & 0\% & 6\% & 94\% \\
Chronological Sequencing & Temp-Dom. & 0\% & 0\% & 100\% \\
\bottomrule
\end{tabular}%
}
\caption{Per-task failure-type breakdown. Fact-dominant tasks fail almost exclusively via knowledge gaps; temporal-dominant tasks fail predominantly via relational reasoning errors.}
\label{tab:failure_pertask}
\end{table}

Fact-dominant tasks produce knowledge-gap failures almost exclusively (75--100\%), while temporal-dominant tasks produce relational-reasoning failures at 37--100\%. Visual grounding failures are a small minority across all tasks. This task-aligned structure supports the interpretation in Section~\ref{sec:spectrum}: the two dominant failure modes are independent and separable, and poor performance on temporal-dominant tasks cannot be explained by missing cultural knowledge.

\subsection{Partial-Credit Evaluation}
\label{sec:partial_credit}

The main results in Table~1 use exact-match accuracy, which assigns zero credit to partially correct answers (e.g., a chronological ordering that misplaces one of four artifacts). To test whether this strictness materially affects our conclusions, we re-scored the model outputs using partial-credit metrics appropriate to each task: Kendall's $\tau$ for Chronological Sequencing, multi-label F1 for the multi-select tasks (Material Availability, Manufacturing Technique), and Jaccard similarity for Period Grouping. Results are reported from a single run.

\begin{table}[h!]
\centering
\small
\resizebox{\columnwidth}{!}{%
\begin{tabular}{lcccc}
\toprule
\textbf{Model} & \textbf{Chron.} & \textbf{Mat.\ Avail.} & \textbf{Mfg.\ Tech.} & \textbf{Period Group.} \\
 & Exact / $\tau$ & Exact / F1 & Exact / F1 & Exact / Jaccard \\
\midrule
GPT-4o & 30.4 / 0.620 & 85.6 / 0.945 & 48.9 / 0.810 & 32.3 / 0.564 \\
Qwen2.5-VL-7B & 14.0 / 0.407 & 69.0 / 0.935 & 37.6 / 0.750 & 21.0 / 0.495 \\
\bottomrule
\end{tabular}%
}
\caption{Exact-match vs.\ partial-credit metrics. Values in percent except $\tau$ and Jaccard, which are in $[0,1]$.}
\label{tab:partial_credit}
\end{table}

Although partial-credit scores are higher in absolute terms, the relative ranking among models and the core conclusion (significantly deficient temporal reasoning on temporal-dominant tasks) remain unchanged. We retain exact-match as the primary metric in the main paper because cultural-heritage applications demand full correctness: a partially correct chronological ordering is still a misrepresentation of the historical record.

\subsection{Reliability of the Automatic Evaluator}
\label{sec:human_eval}

All results in Table~1 are produced by an automatic evaluation pipeline that parses model outputs and checks them against ground-truth answers. To assess the reliability of this pipeline, we conducted a human-evaluation study. Two authors of this paper independently annotated a random 10\% sample of all model responses, labelling each response as either correctly or incorrectly scored by the pipeline. We then computed pairwise Cohen's Kappa ($\kappa$) between all three raters: the two human annotators and the automatic evaluator.

\begin{table}[h!]
\centering
\small
\begin{tabular}{lc}
\toprule
\textbf{Rater Pair} & \textbf{Cohen's} $\kappa$ \\
\midrule
Human 1 $\leftrightarrow$ Human 2 & 0.92 \\
Human 1 $\leftrightarrow$ Automatic Evaluator & 0.85 \\
Human 2 $\leftrightarrow$ Automatic Evaluator & 0.84 \\
\midrule
Average Pairwise $\kappa$ & 0.87 \\
\bottomrule
\end{tabular}
\caption{Pairwise Cohen's Kappa between two human annotators and the automatic evaluator.}
\label{tab:kappa}
\end{table}

Human--human agreement ($\kappa = 0.92$) establishes an upper bound on inter-rater reliability for this task. The automatic evaluator achieves $\kappa = 0.85$ and $\kappa = 0.84$ against the two human annotators, approaching this ceiling. Under the standard interpretation guidelines of Landis and Koch~\cite{landis1977measurement}, all pairwise values fall in the ``almost perfect'' agreement range ($\kappa > 0.8$), indicating that the pipeline performs near human-level consistency in judging correctness.

\subsection{Temporal Imbalance and Textual Metadata}
\label{sec:imbalance_metadata}

The artifact counts in TAB-VLM vary across historical periods (Table~\ref{tab:period_distribution}), and our main evaluation uses visual input only. We examine both factors as potential confounds.

\paragraph{Temporal imbalance.} We computed per-period accuracy on Style-Period Attribution for GPT-4o and Qwen2.5-VL-7B to test whether the number of artifacts per period predicts accuracy on that period.

\begin{table}[h!]
\centering
\small
\resizebox{\linewidth}{!}{
\begin{tabular}{lccc}
\toprule
\textbf{Period} & \textbf{Count} & \textbf{GPT-4o} & \textbf{Qwen2.5-VL-7B} \\
\midrule
Classical (200 BCE -- 650 AD) & 21 & 76.9 & 57.1 \\
Prehistoric (Before 3300 BCE) & 21 & 75.0 & 66.7 \\
Late Medieval (1200 -- 1526 AD) & 17 & 61.5 & 23.5 \\
Bronze Age (3300 -- 1800 BCE) & 13 & 38.5 & 53.8 \\
Iron Age (1800 -- 200 BCE) & 9 & 0.0 & 22.2 \\
Early Modern (1526 -- 1857 AD) & 7 & 66.7 & 57.1 \\
Modern India (1857 -- Present) & 6 & 66.7 & 66.7 \\
Early Medieval (650 -- 1200 AD) & 6 & 0.0 & 33.3 \\
\midrule
Pearson $r$ (count vs.\ acc.) & -- & \multicolumn{2}{c}{0.545 / 0.172} \\
$p$-value & -- & \multicolumn{2}{c}{0.162 / 0.684} \\
\bottomrule
\end{tabular}
}
\caption{Per-period accuracy (\%) on Style-Period Attribution, and Pearson correlation between artifact count and accuracy. Correlations are weak and non-significant.}
\label{tab:temporal_imbalance}
\end{table}

Correlations between artifact count and accuracy are weak and non-significant for both models ($p > 0.1$). The two hardest periods for GPT-4o (Iron Age and Early Medieval, both 0\%) are mid-sized rather than the smallest, ruling out a frequency-driven explanation for the observed accuracy pattern.

\paragraph{Textual metadata.} In realistic cultural-heritage deployments, artifact images are often accompanied by catalog metadata. To test whether our visual-only protocol understates model capability, we evaluated Qwen2.5-VL-7B on a 120-question sample (20 per task) with catalog metadata (object type and main material, excluding any period or dynasty information) appended to the visual prompt.

\begin{table}[h!]
\centering
\small
\resizebox{\linewidth}{!}{
\begin{tabular}{lccc}
\toprule
\textbf{Model} & \textbf{Visual-Only} & \textbf{Visual+Metadata} & \textbf{$\Delta$} \\
\midrule
Qwen2.5-VL-7B & 40.1 & 40.8 & +1.75\% \\
\bottomrule
\end{tabular}
}
\caption{Effect of catalog metadata (object type and material) on accuracy (\%). The gain is minimal, supporting the claim that the bottleneck is temporal reasoning rather than missing input context.}
\label{tab:metadata}
\end{table}

The gain of +1.75\% is small relative to the performance gaps on temporal-dominant tasks. This is consistent with the interpretation in Section~\ref{sec:spectrum} and Section~\ref{sec:failure_modes}: the primary bottleneck is relational temporal reasoning rather than missing auxiliary input.

\subsection{Cross-Cultural Pilot: Western Artifacts}
\label{sec:western_pilot}

The main benchmark focuses on Indian cultural heritage. To assess whether the observed patterns generalise beyond the Indian case and to provide preliminary evidence of a performance gap between Western and non-Western artifacts, we conducted a small pilot study using 84 Western artifacts (Greek/Roman, Medieval, Renaissance, Modern European) sourced from the Cleveland Museum of Art open-access collection. We evaluated GPT-4o on two representative tasks: Style-Period Attribution and Material Availability.

\begin{table}[h!]
\centering
\small
\resizebox{\linewidth}{!}{
\begin{tabular}{lcc}
\toprule
\textbf{Artifact Source} & \textbf{Style-Period (\%)} & \textbf{Material Avail.\ (\%)} \\
\midrule
Indian (TAB-VLM) & 60.3 & 85.3 \\
Western (Pilot) & 67.9 & 97.6 \\
$\Delta$ & +7.6 & +12.3 \\
\bottomrule
\end{tabular}
}
\caption{GPT-4o accuracy on Style-Period Attribution and Material Availability for Indian artifacts (TAB-VLM) vs.\ a Western-artifact pilot (84 items, Cleveland Museum of Art). GPT-4o performs better on Western artifacts on both tasks.}
\label{tab:western_pilot}
\end{table}

GPT-4o scores +7.6 percentage points higher on Style-Period Attribution and +12.3 percentage points higher on Material Availability for Western artifacts relative to Indian artifacts. This gap is consistent with the hypothesis that non-Western cultural heritage presents additional challenges for current VLMs, likely reflecting the distribution of training data. We emphasise that this is a small pilot on two tasks and one model; a full cross-cultural extension of TAB-VLM to other traditions is left to future work.

\end{document}